\documentclass{article} %
\usepackage{iclr2018_conference,times}
\usepackage{hyperref}
\usepackage{url}

\usepackage{graphicx}
\usepackage{amsmath}
\usepackage{amssymb}
\usepackage{booktabs}
\usepackage{multirow}
\usepackage{makecell}
\usepackage{wrapfig}
\usepackage{subfigure}
\usepackage[font=small]{caption}

\iclrfinalcopy

\title{DiracNets: Training Very Deep Neural Networks Without Skip-Connections}

\author{Sergey Zagoruyko, Nikos Komodakis\\
  Universit\'e Paris-Est, \'Ecole des Ponts ParisTech\\
  Paris, France\\
  \texttt{\{sergey.zagoruyko,nikos.komodakis\}@enpc.fr}
}

\newcommand{\R}{\mathbb{R}}

\begin{document}

\maketitle

\begin{abstract}
  Deep neural networks with skip-connections, such as ResNet, show excellent performance in various image classification benchmarks. It is though observed that the initial motivation behind them - training deeper networks - does not actually hold true, and the benefits come from increased capacity, rather than from depth. Motivated by this, and inspired from ResNet, we propose a simple Dirac weight parameterization, which allows us to train very deep plain networks without explicit skip-connections, and achieve nearly the same performance. This parameterization has a minor computational cost at training time and no cost at all at inference, as both Dirac parameterization and batch normalization can be folded into convolutional filters, so that network becomes a simple chain of convolution-ReLU pairs. We are able to match ResNet-1001 accuracy on CIFAR-10 with 28-layer wider plain DiracNet, and closely match ResNets on ImageNet. Our parameterization also mostly eliminates the need of careful initialization in residual and non-residual networks. The code and models for our experiments are available at \url{https://github.com/szagoruyko/diracnets}
\end{abstract}

\section{Introduction}

There were many attempts of training very deep networks. In image classification, after the success of \cite{AlexNet} with AlexNet (8 layers), the major improvement was brought by \cite{Simonyan15} with VGG~(16-19 layers) and later by~\cite{GoogLeNet} with Inception (22 layers). In recurrent neural networks, LSTM by~\cite{Hochreiter97longshort-term} allowed training deeper networks by introducing gated memory cells, leading to major increase of parameter capacity and network performance. Recently, similar idea was applied to image classification by~\cite{highway} who proposed Highway Networks, later improved by~\cite{he2015deep} with Residual Networks, resulting in simple architecture with skip-connections, which was shown to be generalizable to many other tasks. There were also proposed several other ways of adding skip-connections, such as DenseNet by~\cite{huang2016densely}, which passed all previous activations to each new layer.

Despite the success of ResNet, a number of recent works showed that the original motivation of training deeper networks does not actually hold true, e.g.\ it might just be an ensemble of shallower networks~\cite{DBLP:journals/corr/VeitWB16}, and ResNet widening is more effective that deepening~\cite{Zagoruyko2016WRN}, meaning that there is no benefit from increasing depth to more than 50 layers. It is also known that deeper networks can be more efficient than shallower and wider, so various methods were proposed to train deeper networks, such as well-designed initialization strategies and special nonlinearities~\cite{GlorotAISTATS2010,journals/corr/HeZR015, ELU}, additional mid-network losses~\cite{dsn}, better optimizers~\cite{SutskeverMartensDahlHinton_icml2013}, knowledge transfer~\cite{Romero-et-al-TR2014,Chen:ICLR16} and layer-wise training~\cite{Schmidhuber:92ncchunker}.

To summarize, deep networks with skip-connections have the following problems:
\begin{itemize}
  \setlength\itemsep{0.1em}
  \item Feature reuse problem: upper layers might not learn useful representations given previous activations;
  \item Widening is more effective than deepening: there is no benefit from increasing depth;
  \item Actual depth is not clear: it might be determined by the shortest path.
\end{itemize}

However, the features learned by such networks are generic, and they are able to train with massive number of parameters without negative effects of overfitting. We are thus interested in better understanding of networks with skip-connections, which would allow us to train very deep \textit{plain} (without skip-connections) networks and benefits they could bring, such as higher parameter efficiency, better generalization, and improved computational efficiency.

Motivated by this, we propose a novel weight parameterization for neural networks, which we call Dirac parameterization, applicable to a wide range of network architectures. Furthermore, by use of the above parameterization, we propose novel plain VGG and ResNet-like architecture without explicit skip-connections, which we call DiracNet. These networks are able to train with hundreds of layers, surpass 1001-layer ResNet while having only 28-layers, and approach Wide ResNet (WRN) accuracy. We should note that we train DiracNets end-to-end, without any need of layer-wise pretraining. We believe that our work is an important step towards simpler and more efficient deep neural networks.

Overall, our contributions are the following:

\begin{itemize}
  \setlength\itemsep{0.1em}
  \item We propose generic Dirac weight parameterization, applicable to a wide range of neural network architectures;
  \item Our plain Dirac parameterized networks are able to train end-to-end with hundreds of layers. Furthermore, they are able to train with massive number of parameters and still generalize well without negative effects of overfitting;
  \item Dirac parameterization can be used in combination with explicit skip-connections like ResNet, in which case it eliminates the need of careful initialization.
  \item In a trained network Dirac-parameterized filters can be folded into a single vector, resulting in a simple and easily interpretable VGG-like network, a chain of convolution-ReLU pairs.
\end{itemize}

\section{Dirac parameterization}

Inspired from ResNet, we parameterize weights as a residual of Dirac function, instead of adding explicit skip connection. Because convolving any input with Dirac results in the same input, this helps propagate information deeper in the network. Similarly, on backpropagation it helps alleviate vanishing gradients problem.

\newcommand{\vect}[1]{\mathbf{#1}}
\newcommand{\vectm}[1]{\(\vect{#1}\)}
\newcommand{\deltadef}{\vect{I}}

Let \vectm{I} be the identity in algebra of discrete convolutional operators, i.e.\ convolving it with input \vectm{x} results in the same output \vectm{x} (\(\odot\) denotes convolution):

\begin{equation}
  \deltadef \odot \vect{x} = \vect{x}
  \label{eq:deltax=x}
\end{equation}

In two-dimensional case convolution might be expressed as matrix multiplication, so \vectm{I} is simply an identity matrix, or a Kronecker delta \(\delta\). We generalize this operator to the case of a convolutional layer, where input $\vect{x} \in \R^{M,N_1,N_2,\ldots,N_L}$ (that consists of $M$ channels of spatial dimensions ($N_1$, $N_2$, ..., $N_L$)) is convolved with weight $\vect{\hat{W}}\in\R^{M,M,K_1,K_2,\ldots,K_L}$ (combining M filters\footnote{outputs are over the first dimension of $\vect{\hat{W}}$,
inputs are over the second dimension of $\vect{\hat{W}}$}) to produce an output \vectm{y} of $M$ channels, i.e. $\vect{y} = \vect{\hat{W}}\odot \vect{x}$.
In this case we define Dirac delta $\deltadef \in \R^{M,M,K_1,K_2,\ldots,K_L}$, preserving eq.~(\ref{eq:deltax=x}), as the following:
\begin{equation}
  \deltadef(i,j,l_1,l_2,\ldots,l_L) =
  \begin{cases}
    1 & \text{if $i=j$ and $l_m \leq K_m $ for $m=1..L$},\\
    0 & \text{otherwise};
  \end{cases}
\end{equation}

Given the above definition, for a convolutional layer \(\vect{y}=\vect{\hat{W}}\odot \vect{x}\) we propose the following parameterization for the weight \(\vect{\hat{W}}\) (hereafter we omit bias for simplicity):

\newcommand{\diag}[1]{\mathrm{diag}(#1)}
\begin{align}\label{eq:main}
  \vect{y} &= \vect{\hat{W}} \odot \vect{x},\\
  \vect{\hat{W}} &= \diag{\vect{a}}\deltadef + \vect{W},
\end{align}
where \(\vect{a} \in \R^M\) is scaling vector learned during training, and \vectm{W} is a weight vector. Each \(i\)-th element of \vectm{a} corresponds to scaling of \(i\)-th filter of \vectm{W}.  When all elements of \vectm{a} are close to zero, it reduces to a simple linear layer \(\vect{W} \odot \vect{x}\). When they are higher than 1 and \vectm{W} is small, Dirac dominates, and the output is close to be the same as input.

We also use weight normalization \cite{Salimans2016WeightNorm} for \vectm{W}, which we find useful for stabilizing training of very deep networks with more than 30 layers:
\begin{equation}
  \vect{\hat{W}} = \diag{\vect{a}}\deltadef + \diag{\vect{b}}\vect{W}_{\mathrm{norm}},
  \label{eq:weightnorm}
\end{equation}
where \(\vect{b} \in \R^M\) is another scaling vector (to be learned during training), and \(\vect{W}_{\mathrm{norm}}\) is a normalized weight vector where each filter is normalized by it's Euclidean norm.
We initialize \vectm{a} to 1.0 and \vectm{b} to 0.1, and do not \(l_2\)-regularize them during training, as it would lead to degenerate solutions when their values are close to zero. We initialize \vectm{W} from normal distribution \(\mathcal{N}(0, 1)\). Gradients of (\ref{eq:weightnorm}) can be easily calculated via chain-rule. We rely on automatic differentiation, available in all major modern deep learning frameworks (PyTorch, Tensorflow, Theano), to implement it.

Overall, this adds a negligible number of parameters to the network (just two scaling multipliers per channel) during training, which can be folded into filters at test time. %

\vspace{0.5cm}

\subsection{Connection to ResNet}

Let us discuss the connection of Dirac parameterization to ResNet. Due to distributivity of convolution, eq.~(\ref{eq:main}) can be rewritten to show that the skip-connection in Dirac parameterization is implicit (we omit \vectm{a} for simplicity):
\begin{equation}
  \vect{y} = \sigma\big((\vect{I} + \vect{W}) \odot \vect{x}\big) = \sigma\big(\vect{x} + \vect{W} \odot \vect{x}\big),
  \label{eq:resnet}
\end{equation}
where $\sigma(x)$ is a function combining nonlinearity and batch normalization. The skip connection in ResNet is explicit:
\begin{equation}
  \vect{y} = \vect{x} + \sigma (\vect{W} \odot \vect{x})
\end{equation}

This means that Dirac parameterization and ResNet differ only by the order of nonlinearities. Each delta parameterized layer adds complexity by having unavoidable nonlinearity, which is not the case for ResNet. Additionally, Dirac parameterization can be folded into a single weight vector on inference.

\section{Experimental results}

We adopt architecture similar to ResNet and VGG, and instead of explicit skip-connections use Dirac parameterization (see table~\ref{table:arch}). The architecture consists of three groups, where each group has $2N$ convolutional layers ($2N$ is used for easier comparison with basic-block ResNet and WRN, which have $N$ blocks of pairs of convolutional layers per group). For simplicity we use max-pooling between groups to reduce spatial resolution. We also define width $k$ as in WRN to control number of parameters.

We chose CIFAR and ImageNet for our experiments. As for baselines, we chose Wide ResNet with identity mapping in residual block \cite{basicblock2} and basic block (two $3\times3$ convolutions per block). We used the same training hyperparameters as WRN for both CIFAR and ImageNet.

The experimental section is composed as follows. First, we provide a detailed experimental comparison between plain and plain-Dirac networks, and compare them with ResNet and WRN on CIFAR.
Also, we analyze evolution of scaling coefficients during training and their final values. Then, we present ImageNet results.
Lastly, we apply Dirac parameterization to ResNet and show that it eliminates the need of careful initialization.

\newcommand{\blocka}[2]{
\(
\begin{bmatrix}
  \text{3$\times$3, #1}\\[-.1em]
\end{bmatrix}
\)$\times$2#2
}
\newcommand{\convsize}[1]{#1$\times$#1}
\newcommand{\convname}[1]{\texttt{#1}}
\def\cellheight{0.34cm}
\begin{table}[ht]
  \centering
  \begin{tabular}{ccc}
    \toprule
    name & output size & layer type \\
    \midrule
    \texttt{conv1} & $32\times32$ & [3$\times$3, 16] \\
    \convname{group1} & \convsize{32} & \blocka{$16\times16k$}{$N$} \\
    \texttt{max-pool} & \convsize{16} & \\
    \convname{group2} & \convsize{16} & \blocka{$32k\times32k$}{$N$} \\
    \texttt{max-pool} & \convsize{8} & \\
    \convname{group3} & \convsize{8} & \blocka{$64k\times64k$}{$N$} \\
    \texttt{avg-pool} & $1\times1$ & [$8\times8$] \\
    \bottomrule
  \end{tabular}
  \caption{Structure of DiracNets. Network width is determined by factor $k$. Groups of convolutions are shown in brackets as [kernel shape, number of input channels, number of output channels] where  $2N$ is a number of layers in a group. Final classification layer and dimensionality changing layers are omitted for clearance.}
  \label{table:arch}
\end{table}

\subsection{Plain networks with Dirac parameterization}

\begin{figure}[ht]
  \centering
  \includegraphics[scale=0.7]{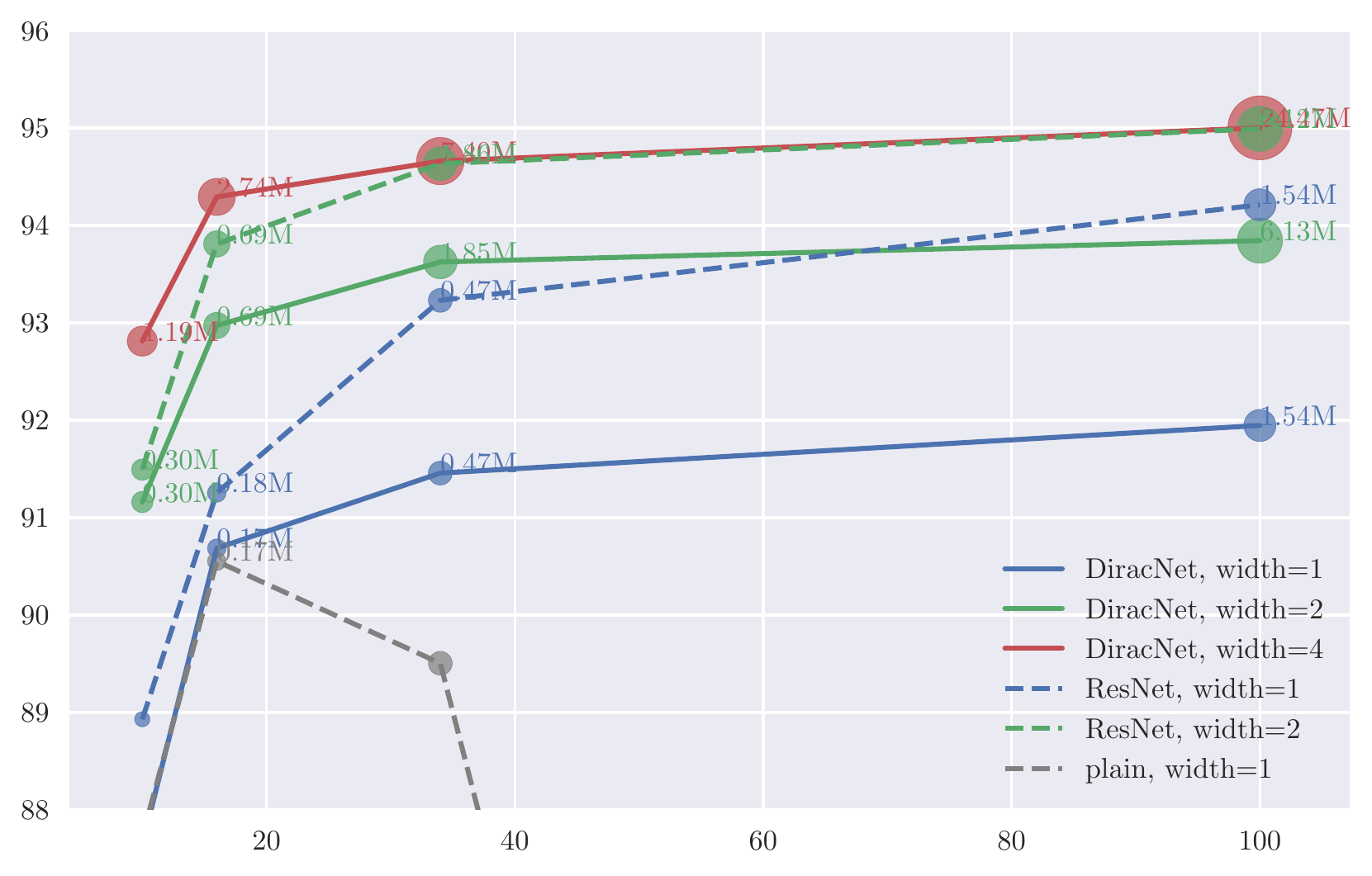}
  \caption{DiracNet and ResNet with different depth/width, each circle area is proportional to number of parameters.
  DiracNet needs more width (i.e.\ parameters) to match ResNet accuracy. Accuracy is calculated as median of 5 runs.}
  \label{fig:param_circles}
\end{figure}

In this section we compare plain networks with plain DiracNets. To do that, we trained both with 10-52 layers and the same number of parameters at the same depth (fig.~\ref{fig:param_circles}).
As expected, at 10 and 16 layers there is no difficulty in training plain networks, and both plain and plain-Dirac networks achieve the same accuracy.
After that, accuracy of plain networks quickly drops, and with 52 layers only achieves 88\%, whereas for Dirac parameterized networks it keeps growing.
DiracNet with 34 layers achieves 92.8\% validation accuracy, whereas simple plain only 91.2\%.
Plain 100-layer network does not converge and only achieves 40\% train/validation accuracy, whereas DiracNet achieves 92.4\% validation accuracy.

\subsection{Plain Dirac networks and residual networks}

To compare plain Dirac parameterized networks with WRN we trained them with different width \(k\) from 1 to 4 and depth from 10 to 100 (fig.~\ref{fig:param_circles}). As observed by WRN authors, accuracy of ResNet is mainly determined by the number of parameters, and we even notice that wider networks achieve better performance than deeper. DiracNets, however, benefit from depth, and deeper networks with the same accuracy as wider have less parameters. In general, DiracNets need more parameters than ResNet to achieve top accuracy, and we were able to achieve 95.25\% accuracy with DiracNet-28-10 with 36.5M parameters, which is close to WRN-28-10 with 96.0\% and 36.5M parameters as well. We do not observe validation accuracy degradation when increasing width, the networks still perform well despite the massive number of parameters, just like WRN. Interestingly, plain DiracNet with only 28 layers is able to closely match ResNet with 1001 layers (table~\ref{table:main})

\begin{table}
  \centering
  \begin{tabular}{lcccc}
    \toprule
     & depth-width & \# params & CIFAR-10 & CIFAR-100 \\ \midrule
     NIN, \cite{nin} & & & 8.81 & 35.67 \\
     ELU, \cite{ELU} & & & 6.55 & 24.28 \\
     VGG  & 16 & 20M & 6.09$\pm$0.11 & 25.92$\pm$0.09 \\
     \multirow{2}{*}{DiracNet (ours)} & 28-5 & 9.1M & 5.16$\pm$0.14 & 23.44$\pm$0.14 \\
                                      & 28-10 & 36.5M & 4.75$\pm$0.16 & 21.54$\pm$0.18 \\
     \midrule
     ResNet & 1001-1 & 10.2M & 4.92 & 22.71 \\
     Wide ResNet & 28-10 & 36.5M & \textbf{4.00} & \textbf{19.25} \\
     \bottomrule
  \end{tabular}
  \caption{CIFAR performance of plain (top part) and residual (bottom part) networks on with horizontal flips and crops data augmentation. DiracNets outperform all other plain networks by a large margin, and approach residual architectures. No dropout it used. For VGG and DiracNets we report mean$\pm$std of 5 runs.}
  \label{table:main}
\end{table}

\subsection{Analysis of scaling coefficients}

\begin{figure}
  \centering
  \includegraphics[width=\textwidth]{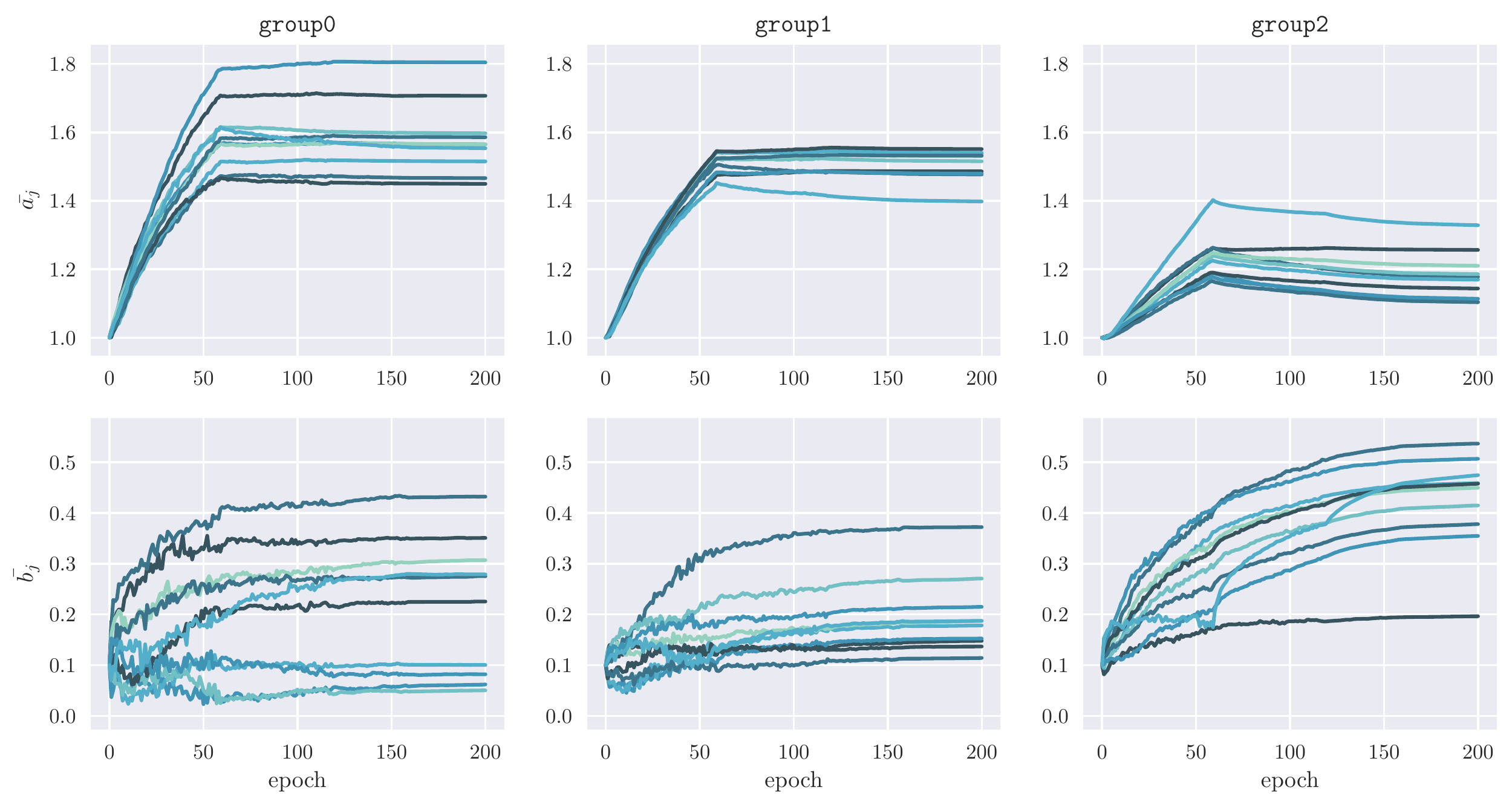}
  \caption{Average values of \vectm{a} and \vectm{b} during training for different layers of DiracNet-34. Deeper color means deeper layer in a group of blocks.}
  \label{fig:a_and_b}
\end{figure}

As we leave \vectm{a} and \vectm{b} free of \(l_2\)-regularization, we can visualize significance of various layers and how it changes during training by plotting their averages \vectm{\bar{a}} and \vectm{\bar{b}}, which we did for DiracNet-34 trained on CIFAR-10 on fig.~\ref{fig:a_and_b}. Interestingly, the behaviour changes from lower to higher groups of the network with increasing dimensionality. We also note that no layers exhibit degraded \vectm{a} to \vectm{b} ratio, meaning that all layers are involved in training. We also investigate these ratios in individual feature planes, and find that the number of degraded planes is low too.

\subsection{Dirac parameterization for ResNet weight initialization}

As expected, Dirac parameterization does not bring accuracy improvements to ResNet on CIFAR, but eliminates the need of careful initialization. To test that, instead of usually used MSRA init \cite{journals/corr/HeZR015}, we parameterize weights as:
\begin{equation}
  \vect{\hat{W}} = \vect{I} + \vect{W},
\end{equation}
omitting other terms of eq.~(\ref{eq:weightnorm}) for simplicity, and initialize all weights from a normal distribution $\mathcal{N}(0,\sigma^2)$, ignoring filter shapes. Then, we vary $\sigma$ and observe that ResNet-28 converges to the same validation accuracy with statistically insignificant deviations, even for very small values of $\sigma$ such as $10^{-8}$, and only gives slightly worse results when $\sigma$ is around 1. It does not converge when all weights are zeros, as expected. Additionally, we tried to use the same orthogonal initialization as for DiracNet and vary it's scaling, in which case the range of the scaling gain is even wider.

\subsection{ImageNet results}

\begin{figure}
  \centering
  \includegraphics[width=\textwidth]{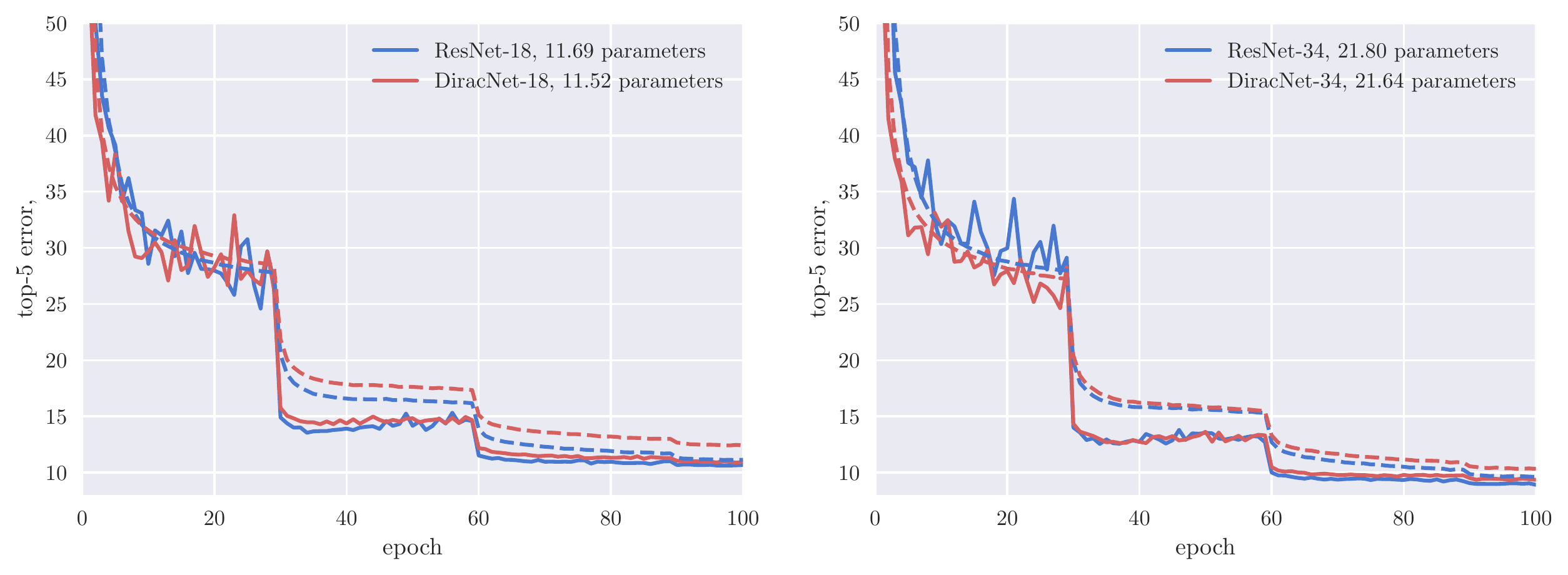}
  \caption{Convergence of DiracNet and ResNet on ImageNet. Training top-5 error is shown with dashed lines, validation - with solid. All networks are trained using the same optimization hyperparameters. DiracNet closely matches ResNet accuracy with the same number of parameters.}
  \label{fig:imagenet_convergence}
\end{figure}

We trained DiracNets with 18 and 34 layers and their ResNet equivalents on ILSVRC2012 image classification dataset. We used the same setup as for ResNet training, and kept the same number of blocks per groups. Unlike on CIFAR, DiracNet almost matches ResNet in accuracy (table~\ref{table:imagenet}), with very similar convergence curves (fig.~\ref{fig:imagenet_convergence}) and the same number of parameters. As for simple plain VGG networks, DiracNets achieve same accuracy with 10 times less parameters, similar to ResNet.

Our ImageNet pretrained models and their simpler folded convolution-ReLU chain variants are available at \url{https://github.com/szagoruyko/diracnets}.

\begin{table}
  \centering
  \begin{tabular}{clccc}
    \toprule
    & Network & \# parameters & top-1 error & top-5 error \\
    \midrule
    \multirow{4}{*}{plain} & VGG-CNN-S \cite{Chatfield14} & 102.9M & 36.94 & 15.40 \\
    & VGG-16 \cite{Simonyan15} & 138.4M & 29.38 & - \\
    & DiracNet-18 & 11.7M & 30.37 & 10.88 \\
    & DiracNet-34 & 21.8M & 27.79 & 9.34 \\
    \midrule
    \multirow{2}{*}{residual} & ResNet-18 [our baseline] & 11.7M & 29.62 & 10.62 \\
    & ResNet-34 [our baseline] & 21.8M & 27.17 & 8.91 \\
    \bottomrule
  \end{tabular}
  \vspace{0.2cm}
  \caption{Single crop top-1 and top-5 error on ILSVRC2012 validation set for plain (top) and residual (bottom) networks.}
  \label{table:imagenet}
\end{table}

\section{Discussion}

We presented Dirac-parameterized networks, a simple and efficient way to train very deep networks with nearly state-of-the-art accuracy. Even though they are able to successfully train with hundreds of layers, after a certain number of layers there seems to be very small or no benefit in terms of accuracy for both ResNets and DiracNets. This is likely caused by underuse of parameters in deeper layers, and both architectures are prone to this issue to a different extent.

Even though on large ImageNet dataset DiracNets are able to closely match ResNet in accuracy with the same number of parameters and a simpler architecture, they are significantly behind on smaller CIFAR datasets, which we think is due to lack of regularization, more important on small amounts of data. Due to use of weight normalization and free scaling parameters DiracNet is less regularized than ResNet, which we plan to investigate in future.

We also observe that DiracNets share the same property as WRN to train with massive number of parameters and still generalize well without negative effects of overfitting, which was initially thought was due to residual connections. We now hypothesize that it is due to a combination of SGD with momentum at high learning rate, which has a lot of noise, and stabilizing factors, such as residual or Dirac parameterization, batch normalization, etc.

\bibliography{egbib}

\begin{thebibliography}{21}
\providecommand{\natexlab}[1]{#1}
\providecommand{\url}[1]{\texttt{#1}}
\expandafter\ifx\csname urlstyle\endcsname\relax
  \providecommand{\doi}[1]{doi: #1}\else
  \providecommand{\doi}{doi: \begingroup \urlstyle{rm}\Url}\fi

\bibitem[Bengio \& Glorot(2010)Bengio and Glorot]{GlorotAISTATS2010}
Yoshua Bengio and Xavier Glorot.
\newblock Understanding the difficulty of training deep feedforward neural
  networks.
\newblock In \emph{Proceedings of AISTATS 2010}, volume~9, pp.\  249--256, May
  2010.

\bibitem[Chatfield et~al.(2014)Chatfield, Simonyan, Vedaldi, and
  Zisserman]{Chatfield14}
K.~Chatfield, K.~Simonyan, A.~Vedaldi, and A.~Zisserman.
\newblock Return of the devil in the details: Delving deep into convolutional
  nets.
\newblock In \emph{British Machine Vision Conference}, 2014.

\bibitem[Chen et~al.(2016)Chen, Goodfellow, and Shlens]{Chen:ICLR16}
T.~Chen, I.~Goodfellow, and J.~Shlens.
\newblock Net2net: Accelerating learning via knowledge transfer.
\newblock In \emph{International Conference on Learning Representation}, 2016.

\bibitem[Clevert et~al.(2015)Clevert, Unterthiner, and Hochreiter]{ELU}
Djork{-}Arn{\'{e}} Clevert, Thomas Unterthiner, and Sepp Hochreiter.
\newblock Fast and accurate deep network learning by exponential linear units
  (elus).
\newblock \emph{CoRR}, abs/1511.07289, 2015.

\bibitem[He et~al.(2015{\natexlab{a}})He, Zhang, Ren, and Sun]{he2015deep}
Kaiming He, Xiangyu Zhang, Shaoqing Ren, and Jian Sun.
\newblock Deep residual learning for image recognition.
\newblock \emph{CoRR}, abs/1512.03385, 2015{\natexlab{a}}.

\bibitem[He et~al.(2015{\natexlab{b}})He, Zhang, Ren, and
  Sun]{journals/corr/HeZR015}
Kaiming He, Xiangyu Zhang, Shaoqing Ren, and Jian Sun.
\newblock Delving deep into rectifiers: Surpassing human-level performance on
  imagenet classification.
\newblock \emph{CoRR}, abs/1502.01852, 2015{\natexlab{b}}.

\bibitem[He et~al.(2016)He, Zhang, Ren, and Sun]{basicblock2}
Kaiming He, Xiangyu Zhang, Shaoqing Ren, and Jian Sun.
\newblock Identity mappings in deep residual networks.
\newblock \emph{CoRR}, abs/1603.05027, 2016.

\bibitem[Hochreiter \& Schmidhuber(1997)Hochreiter and
  Schmidhuber]{Hochreiter97longshort-term}
Sepp Hochreiter and J{\"u}rgen Schmidhuber.
\newblock Long short-term memory, 1997.

\bibitem[Huang et~al.(2016)Huang, Liu, Weinberger, and van~der
  Maaten]{huang2016densely}
Gao Huang, Zhuang Liu, Kilian~Q Weinberger, and Laurens van~der Maaten.
\newblock Densely connected convolutional networks.
\newblock \emph{arXiv preprint arXiv:1608.06993}, 2016.

\bibitem[Krizhevsky et~al.(2012)Krizhevsky, Sutskever, and Hinton]{AlexNet}
A.~Krizhevsky, I.~Sutskever, and G.~Hinton.
\newblock Imagenet classification with deep convolutional neural networks.
\newblock In \emph{NIPS}, 2012.

\bibitem[{Lee} et~al.(2014){Lee}, {Xie}, {Gallagher}, {Zhang}, and {Tu}]{dsn}
C.-Y. {Lee}, S.~{Xie}, P.~{Gallagher}, Z.~{Zhang}, and Z.~{Tu}.
\newblock {Deeply-Supervised Nets}.
\newblock 2014.

\bibitem[Lin et~al.(2013)Lin, Chen, and Yan]{nin}
Min Lin, Qiang Chen, and Shuicheng Yan.
\newblock Network in network.
\newblock \emph{CoRR}, abs/1312.4400, 2013.

\bibitem[Romero et~al.(2014)Romero, Ballas, Ebrahimi~Kahou, Chassang, Gatta,
  and Bengio]{Romero-et-al-TR2014}
Adriana Romero, Nicolas Ballas, Samira Ebrahimi~Kahou, Antoine Chassang, Carlo
  Gatta, and Yoshua Bengio.
\newblock {FitNets}: Hints for thin deep nets.
\newblock Technical Report Arxiv report 1412.6550, arXiv, 2014.

\bibitem[Salimans \& Kingma(2016)Salimans and Kingma]{Salimans2016WeightNorm}
Tim Salimans and Diederik~P. Kingma.
\newblock Weight normalization: A simple reparameterization to accelerate
  training of deep neural networks.
\newblock In \emph{Neural Information Processing Systems 2016}, 2016.

\bibitem[Schmidhuber(1992)]{Schmidhuber:92ncchunker}
J.~Schmidhuber.
\newblock Learning complex, extended sequences using the principle of history
  compression.
\newblock \emph{Neural Computation}, 4\penalty0 (2):\penalty0 234--242, 1992.

\bibitem[Simonyan \& Zisserman(2015)Simonyan and Zisserman]{Simonyan15}
K.~Simonyan and A.~Zisserman.
\newblock Very deep convolutional networks for large-scale image recognition.
\newblock In \emph{ICLR}, 2015.

\bibitem[Srivastava et~al.(2015)Srivastava, Greff, and Schmidhuber]{highway}
Rupesh~Kumar Srivastava, Klaus Greff, and J{\"{u}}rgen Schmidhuber.
\newblock Highway networks.
\newblock \emph{CoRR}, abs/1505.00387, 2015.

\bibitem[Sutskever et~al.(2013)Sutskever, Martens, Dahl, and
  Hinton]{SutskeverMartensDahlHinton_icml2013}
Ilya Sutskever, James Martens, George~E. Dahl, and Geoffrey~E. Hinton.
\newblock On the importance of initialization and momentum in deep learning.
\newblock In Sanjoy Dasgupta and David Mcallester (eds.), \emph{Proceedings of
  the 30th International Conference on Machine Learning (ICML-13)}, volume~28,
  pp.\  1139--1147. JMLR Workshop and Conference Proceedings, May 2013.

\bibitem[Szegedy et~al.(2015)Szegedy, Liu, Jia, Sermanet, Reed, Anguelov,
  Erhan, Vanhoucke, and Rabinovich]{GoogLeNet}
C.~Szegedy, W.~Liu, Y.~Jia, P.~Sermanet, S.~Reed, D.~Anguelov, D.~Erhan,
  V.~Vanhoucke, and A.~Rabinovich.
\newblock Going deeper with convolutions.
\newblock In \emph{CVPR}, 2015.

\bibitem[Veit et~al.(2016)Veit, Wilber, and
  Belongie]{DBLP:journals/corr/VeitWB16}
Andreas Veit, Michael~J. Wilber, and Serge~J. Belongie.
\newblock Residual networks are exponential ensembles of relatively shallow
  networks.
\newblock \emph{CoRR}, abs/1605.06431, 2016.

\bibitem[Zagoruyko \& Komodakis(2016)Zagoruyko and Komodakis]{Zagoruyko2016WRN}
Sergey Zagoruyko and Nikos Komodakis.
\newblock Wide residual networks.
\newblock In \emph{BMVC}, 2016.

\end{thebibliography}
\bibliographystyle{iclr2018_conference}
\end{document}